\documentclass[aps,pre,floats,onecolumn,showpacs,superscriptaddress]{revtex4}

\usepackage{graphicx,epsfig}
\usepackage{times}
\usepackage{graphics,dcolumn,bm,fleqn,epic,eepic,float}
\usepackage{amssymb,amsmath,multirow,rotate,color}

\begin{document}

\title{A Computational Model to Disentangle Semantic Information Embedded in Word Association Norms}

\author{Javier Borge}

\affiliation{Departament d'Enginyeria Inform{\`a}tica i
  Matem{\`a}tiques, Universitat Rovira i Virgili, 43007 Tarragona,
  Spain}

\author{Alex Arenas}

\affiliation{Departament d'Enginyeria Inform{\`a}tica i
  Matem{\`a}tiques, Universitat Rovira i Virgili, 43007 Tarragona,
  Spain}


\begin{abstract}

Two well-known databases of semantic relationships between pairs of words used in psycholinguistics, feature-based and association-based, are studied as complex networks. We propose an algorithm to disentangle feature based relationships from free association semantic networks. The algorithm uses the rich topology of the free association semantic network to produce a new set of relationships between words similar to those observed in feature production norms.

\end{abstract}

\maketitle

\section{Introduction}

Understanding the structure of semantic knowledge is an open challenge of fundamental importance in cognitive science. Along the most powerful computational probabilistic approaches to this challenge \cite{landauer_a,blei,griffiths_pnas,steyvers04,griffiths1}, recent studies have used also the perspective offered by the theory of complex networks to gain insight on it \cite{sigman,steyvers05}. The main idea behind the network approach is to map empirical data onto a graph (usually called complex network) that summarizes the observed relations between words in a given experiment. Once the network is constructed, its statistical characterization (distribution of degree of nodes, clustering measures, etc.) reveal properties that can help to better understand the large-scale structure of semantic relations in the specific set. However, while the network approach has been merely descriptive up to now, computational models like LSA \cite{landauer_a}, WAS \cite{steyvers04} or the Topic Model \cite{griffiths_pnas} have an intrinsic predictive capability. In particular, some of these models are used to reveal interaction between episodic and semantic memory, considering empirical data that reflects the impact of environmental (i.e. nonlinguistic) experience upon linguistic phenomena \cite{steyvers04,andrews05,silberman07}. 

The description of semantic knowledge as a complex network of interactions between words, does not suffice to get a clear picture of the specific relations between complex networks representing different semantic empirical data sets. One of the main reasons for this is that while the notion of node is quite uncontroversial (in our case a word), the concept of edge is not so because it must be committed to a definition of relationship. In what semantics is concerned, we can consider that a word is related to another one if they belong to the same class (category-related, such as \emph{car} and \emph{wagon}); or if they tend to co-occur in many contexts (\emph{car} and \emph{road}); or if they have a cause-effect relationship (\emph{fire} and \emph{smoke}), and so on. For some of these types of relationship there exist empirical data that quantify how strong two words are related. (Notice that two words may have several of these relationships).

It is clear that different semantic networks will arise depending on the type of association used to link words by the subjects of a cognitive experiment. Moreover, given the intricate complexity of human mind, the more free the association scenario, the more rich the types of relationship will appear. These different association scenarios can reflect semantic or episodic memory contents, depending on the experiment. One of the main challenges is to understand the interaction between both memory representations. In \cite{steyvers04} the authors propose the prediction of semantic similarity effects in episodic memory using empirical data. The procedure applied is a modification of  the general LSA scheme, using singular value decomposition and multidimensional scaling over a specific data set \cite{nelson}. The results show the emergence of feature association groups in a multidimensional space known as Word Association Space (WAS).

We will consider the same problem from a complex network perspective adding a different interpretation of the disentanglement process with plausible cognitive implications. In our work, this prediction is reformulated in the following terms: whether is possible to disentangle similarity relationships from general association words networks by the navigation of the semantic network. We address this question assuming that: (i) Each available data set is a partial exposure to semantic knowledge; (ii) Some data sets are more general than others, they grasp the heterogeneity of the semantic knowledge more precisely; and (iii) as a consequence of (ii), some information from a less general data set might be partially implicit in a more general one. We will construct upon these hypothesis  and propose an algorithm that allows the disentanglement of a type of relationship embedded on the structure of a more general association network. In particular, we will focus on two well-known data sets in English: the free-association database constructed by \cite{nelson}, and the semantic feature production norms by \cite{mcrae}.

\subsection{Feature Production Norms}
Feature Production Norms (FP from now on) were produced by McRae et al. by asking subjects to conceptually recognize features when confronted with a certain word. This feature collection is used to build up a vector of characteristics for each word, where each dimension represents a feature.  In particular, participants are presented with a set of concept names and are asked to produce features they think are important for each concept. Each feature stands as a vector component, with a value that represents its production frequency across participants. These norms include 541 living and nonliving thing concepts, for which semantic closeness or {\em similarity} is computed as the cosine (overlap) between pairs of vectors of characteristics. The cosine is obtained as the dot product between two concept vectors, divided by the product of their lengths.

As a consequence, words like \emph{banjo} and \emph{accordion} are very similar (i.e. they have a projection close to 1) because their vector representations show a high overlap, essentially provoked by their shared features as musical instruments, while the vectors for \emph{banjo} and \emph{spider} are very different, showing an overlap close to 0 (orthogonal vectors).\\
In terms of network modeling, each node represents a word, and an edge (or link) is set up between a pair of nodes whenever their vectors  projection is different from 0. The meaning of an edge in this network is thus the features similarity between two words. The network is undirected (symmetric relationships) and weighted by the value of the projections.

\subsection{Free-Association Norms}
Nelson et al. produced these norms by asking over 6000 participants to write down the first word (\emph{target}) that came to their mind when confronted with a \emph{cue} (word presented to the subject). The experiment was performed using more than 5000 cues. Among other information, a frequency of coincidence between subjects for each pair of words is obtained. As an example, words \emph{mice} and \emph{cheese} are neighbors in this database, because a large fraction of the subjects related this target to this cue. Note, however, that the association of these two words is not directly represented by similar features but other relationships (in this case mice eat cheese). The network empirically obtained is directed and weighted, however, for the sake of simplicity we will assume that links are bidirectional. This assumption does not affect our goal because certainly in a free association scenario its interpretation is not difficult, e.g in the case mice eat cheese, we also can interpret the backwards relation cheese is eaten by mice. The weights represent the frequency of association in the sample.

Generally speaking, Free-Association Norms (FA from now on) represent a more complex scenario than FP when considering the semantics of edges. FA is heterogeneous by construction, it may grasp any relation between words e.g. a causal-temporal relation (\emph{fire} and \emph{smoke}), an instrumental relation (\emph{broom} and \emph{floor}) or a conceptual relation (\emph{bus} and \emph{train}), among others. This heterogeneity will be on the basis of our approach because we assume that some similarity information is implicit in FA. The temporal dynamics (growth and change) of this network is not considered in this work (see \cite{steyvers05} for a complex network approach to this problem).\\

Our specific goal is to propose a computational model to extract a FP-like network from the track of a dynamical process upon FA. The idea is to simulate a na\"{\i}ve cognitive navigation on top of a general association semantic network to relate words with a certain similarity, in particular we want to recover features similarities. We schematize this process as uncorrelated \emph{ random walks} from node to node that propagate an inheritance mechanism among words, converging to a feature vectors network. Our intuition about the expected success of our approach relies on two facts: the modular structure of the FA network surely retains significant meta-similitude relationships, and random walks are a the most simple dynamical processes capable of revealing the local neighborhoods of nodes when they persistently get trapped into modules. The inheritance mechanism is a simple reinforcement of similarities within these groups. We call this algorithm the Random Inheritance Model (RIM). 

The results obtained show macro-statistical coincidences (functional form of the distributions and descriptors) between the real and the synthetic FP network, moreover, the model yields also significant success at the microscopic level, i.e. is able to reproduce to a large extent FP empirical relationships. These results support the general hypothesis about implicit entangled information in FA, and also reveals a possible mechanism of navigation to recover feature information in semantic networks. Finally, we compare these results with those obtained using LSA and WAS on the same data sets.

\section{Complex Network Theory in a Nutshell}
Complex networks refers to networks (graphs) whose topological characteristics are \emph{non-trivial} in contrast with simple networks where regularities and symmetries dominate their structure. Complex network are found in the representation of interacting elements, in many fields of science ranging from biology to social sciences \cite{newman_review}. Related to cognitive science, the application of complex networks has been specially successful  to better understand some aspects of brain dynamics in neuroscience \cite{sporns} and linguistics \cite{sigman,motter,bales,mehler}, and there is evidence that a step further to psycholinguistics is also on its way \cite{steyvers05,sole}.

Complex networks main assets comprise a wide range of measures that help on the quantification of its structural characteristics, either at a \emph{micro} (node), \emph{meso} (group) and \emph{macro} (network) level. Here we outline some key concepts on complex networks that will be used or referred in this work. For extensive network theory reviews and foundational works, see \cite{newman_review,watts,barabasi,physrepYamir}.

\subsection{Statistical descriptors of networks}

A graph (or network) is a pair $G = (V, E)$, where $V$ represents the set of \emph{vertices} (or nodes), and $E$ represents the set of \emph{edges} (or links). The \emph{order} of a graph is the number of its vertices, i.e. $N$.
A \emph{directed graph} is a graph with directed edges, commonly named \emph{arcs}. In contrast, when arcs are symmetric, the graph is \emph{undirected}. A \emph{weighted graph} associates a label (weight $w_{ij}$) to every edge in the graph connecting a node $i$ and a node $j$.
Two vertices $v_i$ and $v_j$ are \emph{adjacent}, or neighbors, if they have an edge $(v_i, v_j) \in E$ connecting them. 
A \emph{path} in a graph is a sequence of vertices $v_1, v_2,... v_n$ such that from each of its vertices there is an edge to the next vertex in the sequence $(v_i, v_{i+1}) \in E$ for $1 < i < n$. The first vertex is called the \emph{start} vertex and the last vertex is called the \emph{end} vertex. 
The \emph{length} of the path or \emph{distance} between $v_1$ and $v_n$ i.e. the number of edges in the path. For weighted graphs, the length is the addition of each weight in the path. When $v_1$ and $v_n$ are identical, their distance is 0 by definition. When $v_1$ and $v_n$ are unreachable from each other, their distance is defined to be infinity ($\infty$). Finally, a \emph{connected graph} is an undirected graph such that there exists a path between all pairs of vertices. 

The number of links of a given node (words, in our networks) is called its degree. If we indicate the total number of links by $L$, then the average degree is $\langle k \rangle = \frac{2L}{N}$. The distribution of degrees of a network, $p(k)$, is one of its basic statistical characteristics.The distribution $p(k)$ is the fraction of nodes in the network that have degree $k$, or equivalently, the probability that a node chosen uniformly at random has degree $k$. Usually, the cumulative degree distribution $P(k)=p(k'<k)$ is plotted. 
The definitions of average degree and degree distribution are extended from degree $k_i$ to the strength $s_i=\sum_j w_{ij}$, the average strength is $\langle s \rangle$ and the cumulative strength distribution $P(s)$. Empirical results of networks in different disciplines show that many large networks have heavy tailed degree distributions, in the physics literature these networks are referred as scale-free. Scale-free topologies have a relatively small number of well-connected nodes {\em hubs}, and the distribution of node connectivities -$P(k)$ or $P(s)$- follows a power law (i.e. $P(s)\sim s^{-\gamma}$, being $\gamma$ the scaling exponent) . Most large language data sets Thesaurus, Wordnet, TASA corpus, and others, reveal such scale-free topology \cite{motter,steyvers05,sole}.

Other commonly used statistical descriptors are: The clustering coefficient $C_i$ of a vertex $v_i$ defined as the proportion of links between the vertices within its neighborhood divided by the number of links that could possibly exist between them \cite{watts}. Its equation for undirected graphs is:
\begin{equation}
  C_i=\frac{2E_i}{k(k-1)}
\end{equation}
where $E_i$ are the actual edges of vertex $i$. Usually the average clustering coefficient $C=<C_i>_{i}$ of the network is presented. The average path length ($L$) of the network, is defined as the average of the geodesic paths (minimal distance) between any pair of nodes. In the current work we will use also the {\em Diameter} ($D$) of the network referring to the longest distance in the network between any two vertices. The {\em assortativity} $r$, is a measure that indicates the preference for high-degree
vertices to attach to other high-degree vertices, while others show disassortative mixing (high-degree vertices attach to low-degree ones). It is defined as the Pearson correlation coefficient of the degrees at either ends of an edge, and therefore it lies in the range $0 < r < 1$. If $r > 0$ the network is assortative \cite{newman_assort}; if $r < 0$, the network is disassortative; for $r = 0$ there are no correlation between vertex degrees.

\section{The Random Inheritance Model (RIM)}

FA and FP norms can be represented as semantic networks of words, which in turn can be analyzed in terms of descriptors presented in the previous section. Both empirical networks are topologically different, that is, the statistical local and global properties differ significantly from each other. The main differences are concerned to the sparsity of FA, in contrast to the strong density of FP.
Since our goal is to compare a synthetic network obtained from FA, to FP up to a microscopic level, we need both networks to have the same nodes (words). To this end, we have extracted from the databases those words that appear in both, which has left two subnetworks of 376 nodes each. The statistical characteristics of the extracted subnetworks do not differ very much from their complete versions but they do between them, see Table \ref{tab:original_descriptors}. From now on, the subnetworks of 376 words common in FA and FP will be used for comparison purposes.\\

\begin{table}[ht]
\caption{Main statistical descriptors of the networks FA and FP data, and their respective common words' subnetworks. $N$ is the number of nodes; $<s>$ is the average strength; $L$ is the average shortest path length; $D$ is the diameter of the network; $C$ is clustering coefficient and $r$ is the assortativity coefficient.}
\begin{tabular}{c|cc|cc}
Descriptor&FA (complete)&FP (complete)&FA (subnetwork)&FP (subnetwork)\\\hline
$N$&5018&541&376&376\\
$\langle s \rangle$&0.77&20.20&0.26&13.43\\
$L$&3.04&1.68&4.41&1.68\\
$D$&5&5&9&3\\
$C$&0.1862&0.6344&0.1926&0.6253\\
$r$&0.097&0.2609&0.3258&0.2951
\end{tabular}
\label{tab:original_descriptors}
\end{table}

Finally, it is worthwhile to mention the fact that the databases, although they both belong to the psycholinguistic field, they were created in different places and years (affecting the use of language); a different number of subjects were used to build up the norms (affecting the robustness of data), etc. Even the intention (i.e. the type of problem they seek to tackle) of the collections is different. It is important to realize about all these facts in order to understand the amount of uncertainty any model faces when trying to reproduce a particular empirical dataset.

Keeping in mind all these general considerations we can move on to specify how our model works. In what follows, we first specify the logic behind our proposal and, after, we describe the mathematical framework that unifies the different steps. The main logic stages in RIM are:

\begin{enumerate}

\item{ Initialization}:
 
First, every word in the FA network is tagged with an initial vector of characteristics. To avoid initial bias, we choose the vectors to be orthogonal in the canonical basis. That means that every word has associated a vector of $N$-dimensions, being $N$ the number of words in the network, with a component at 1 and the rest at zero.

\item  { Navigation and Inheritance}:

Then a random walk of $S$ steps\footnote{A random walk is a time-reversible finite Markov chain, see \cite{lovasz} for a survey on the topic.} starting at a node $i$ is performed. At every step of the walk, we propose an inheritance mechanism that changes $v_i$ (the initial vector of the word $i$) depending on the visited nodes. Let $s = {s_1, s_2, ..., s_n}$ the set of visited nodes. Then the new vector for node $i$ is computed as: 
\begin{equation}
v_{i} := v_{i} + \displaystyle\sum_{j=1}^n v_{s_j}
\end{equation}

The process is performed in parallel, i.e. the update of the feature vectors is done after completion of the inheritance for every word. At the end, we have a synthetic vector of features for every word in the network.

\item  { Averaging}:

Once the feature vectors have been computed, we build up a synthetic feature similarity network. The network is the result of projecting all pairs of vectors and 
prescribing a weighted link between two words according to this projection. The whole process is iterated (by simulating several runs)  up to convergence of the average of the synthetic feature similarity networks generated at each run. The average, after convergence, is the synthetic feature similarity network we compare with FP.

\end{enumerate}
\vspace{0.5cm}

This algorithm can be algebraically described in terms of Markov chains. Let us define the transition probability of the FA network. The elements of FA ($a_{ij}$) correspond to frequency of first association reported by the participants of the experiments. However, note that the 5018 words that appear on the data set are not all the words that appeared at the experiment, but only those that where at the same time cues in the experiment. That means that the data have to be normalized before having a transition probability matrix. We define the transition probability matrix $P$ as:

\begin{equation}
P_{ij}=\frac{a_{ij}}{\sum_j a_{ij}}
\end{equation}

Note that this matrix is asymmetric, as well as the original matrix FA. We maintain this asymmetry property in our approach to preserve the meaning of the empirical data. Once the matrix $P$ is constructed, the random walkers of different lengths are simply represented by powers of $P$. For example, if we perform random walks of length $2$, after averaging over many realizations we will converge to the transition matrix $P^{2}$, every element $(P^2)_{ij}$ represents the probability of reaching $j$, from $i$, in 2 steps, and the same applies to other length values. The inheritance process proposed, corresponds, in this scenario, to a change of basis, from the canonical basis of the N-dimensional space, to the new basis in the space of transitions $T$  

\begin{equation}
T= \lim_{S\to\infty} \sum_{i=1}^{S} P^{i}
\label{tran}
\end{equation}

The convergence of Eq.(\ref{tran}) is guaranteed by the Perron-Frobenius theorem. In practice, the summation in Eq.(\ref{tran}) converges very fast, limiting the dependence on indirect associative strengths \cite{nelson00}. We tested the behavior up to S=10, although with S=4 we already achieve convergence in T up to $10^{-4}$ in terms of the Hamming distance. All the results for RIM will be expressed for $S=4$ from now on. Finally, the matrix that will represent in our model the feature similarity network (synthetic FP), where similarity is calculated as the cosine of the vectors in the new space, is given by the scalar product of the matrix and its transpose, FS $= TT^{\dagger}$.

\section{Results of RIM}

In this section we present the performance of the RIM in disentangling a Feature Production Norm from the empirical FA. To compare the results with the empirical FP we define a set of measures that can be classified in macroscopic and microscopic similarities. To evaluate macroscopic similarities we will use basic descriptors of complex networks, the strength distribution $P(s)$, along with average (global) quantities already computed in Table \ref{tab:original_descriptors}. To evaluate the microscopic similarities we will compute the success rate on the local structure of the neighborhood of words in both real and synthetic networks. We will also compare the results of our model with those obtained using the well known Latent Semantic Analysis (LSA; \cite{landauer_a,landauer_b}) and Word Association Space (WAS; \cite{steyvers04}). Although LSA's applicability goes beyond the scope of this work, it stands as an appropriate benchmark model to compare the performance of our proposal. In particular, we have used LSA vector representation based on the corpus TASA for the subset of common words in FA and FP, with a space dimensionality of $d=300$. This LSA TASA-based representation is suitable for comparison because it has been assessed as a simulation of human vocabulary test synonym judgments \cite{landauer_1998}. WAS model is specially pertinent for the current comparison because: on one hand, the model is formally similar to LSA; on the other, it makes use of mediated strength between non-direct associates as in RIM, and has been reported on the same data set we use in order to extract semantic information. Accordingly, we have performed the procedure described in the cited article upon the whole network, and after extracted the mentioned 376 subset of words. We only compare to the best results of WAS for this data set (which correspond to Singular Value Decomposition under $S^(2)$, $N=5018$, and $d=400$, see \cite{steyvers04} for details).

\begin{figure}[t]
\centering 
\includegraphics[width=0.75\textwidth,angle=0,clip=] {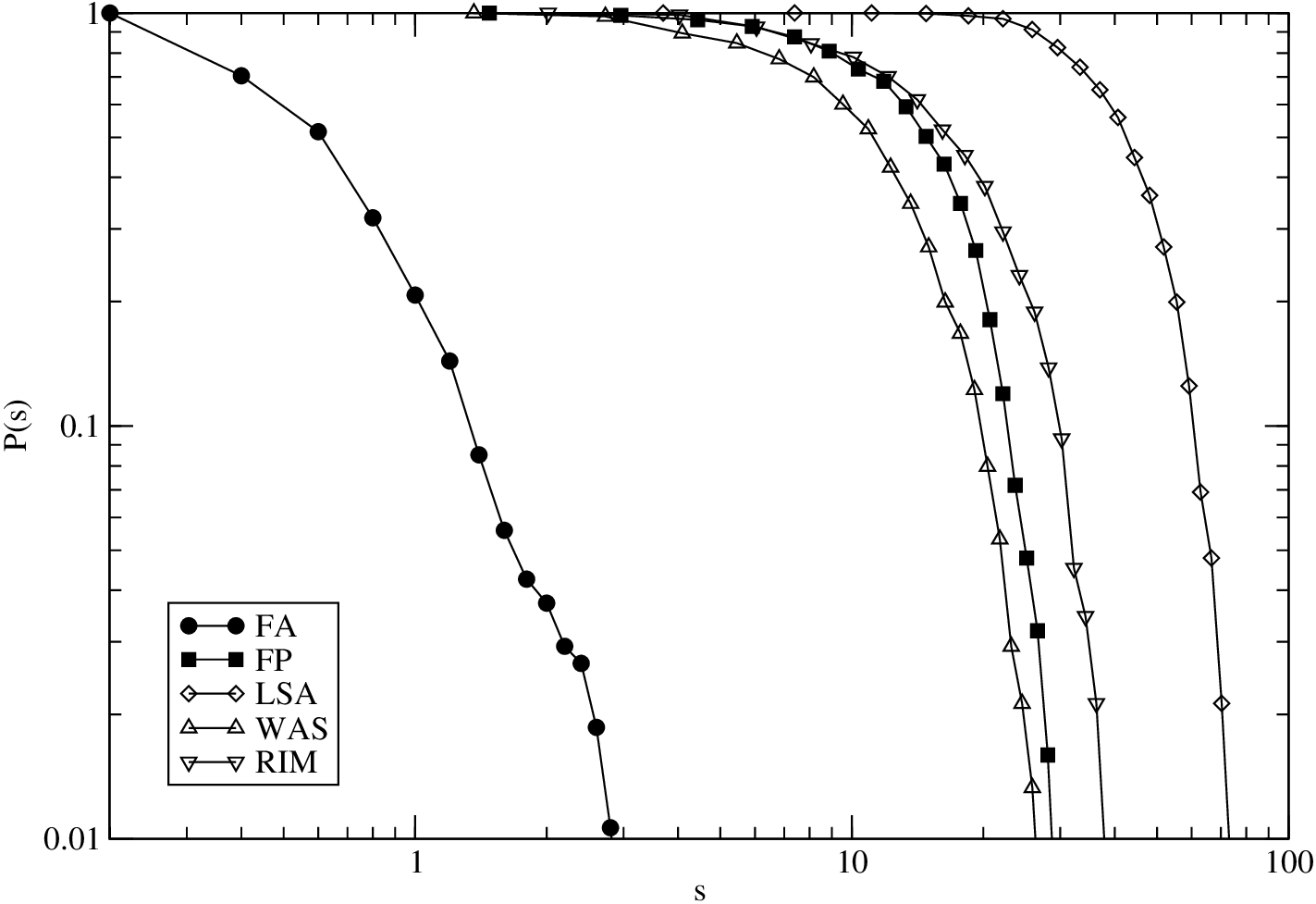}
\caption{Log-log plot of the cumulative strength distribution of the networks: Free Association norms FA (substrate of the dynamic process), Feature Production norms FP (empirical target) , and the synthetic networks obtained using Latent Semantic Analysis LSA and Random Inheritance Model RIM.}
\label{fig:strength}
\end{figure}

\begin{table}[t]
\caption{Statistical parameters for Free Association norms FA (substrate of the dynamic process), Feature Production norms FP (empirical target) , and the synthetic networks obtained using Latent Semantic Analysis LSA, Word Association Space WAS and Random Inheritance Model RIM.}
\begin{tabular}{c|cc|ccc}
Descriptor&FA&FP&LSA&WAS&RIM\\\hline
$N$&376&376&376&376&376\\
$\langle s \rangle$&0.26&13.43&39.60&10.29&15.62\\
$L$&4.41&1.68&0.02&2.00&1.77\\
$D$&9&3&2&4&3\\
$C$&0.1926&0.6253&0.9611&0.4927&0.5848\\
$r$&0.3258&0.2951&0.1254&0.3031&0.3057
\end{tabular}
\label{tab:synthetic_descriptors}
\end{table}

\subsection{Macroscopic similarities}
First we plot the cumulative strength distribution of the empirical networks FA, FP, and the respective synthetic networks provided by LSA or by applying WAS and RIM to FA, see Figure \ref{fig:strength}. The statistical agreement between FP and RIM and WAS is remarkable. The general observation is that all distributions present an exponential decay instead of a power-law decay, being the cutoff of the distribution in FA more pronounced due to its original sparseness. This specific form of the distributions is characteristic of random homogeneous networks.
In Table \ref{tab:synthetic_descriptors} we present the main descriptors of the four networks for comparison purposes. Again, the agreement between the empirical FP and RIM is marked, RIM reproduces with significant accuracy the average strength, the average path length, diameter, clustering and assortativity, of the FP target network. WAS also succeeds largely on the determination of macroscopic properties of the network, while LSA can not be so similar.

\subsection{Microscopic similarities}

The statistical comparison presented before is informative and important, but not definitive to state the validity of the model to disentangle actual information in the original FA network. The difference between our particular network, and general examples used in complex networks theory is that nodes are tagged and then not interchangeable. The specific neighborhood of every word matters, because it reveals semantic relations, and then the degree or the clustering become less relevant than the specific list of neighbors synthetically obtained.
To this end, we pursue the evaluation of RIM in predicting the specific words of each node's neighborhood. 

\begin{figure}[b]
\centering
\includegraphics[width=0.75\textwidth,clip=] {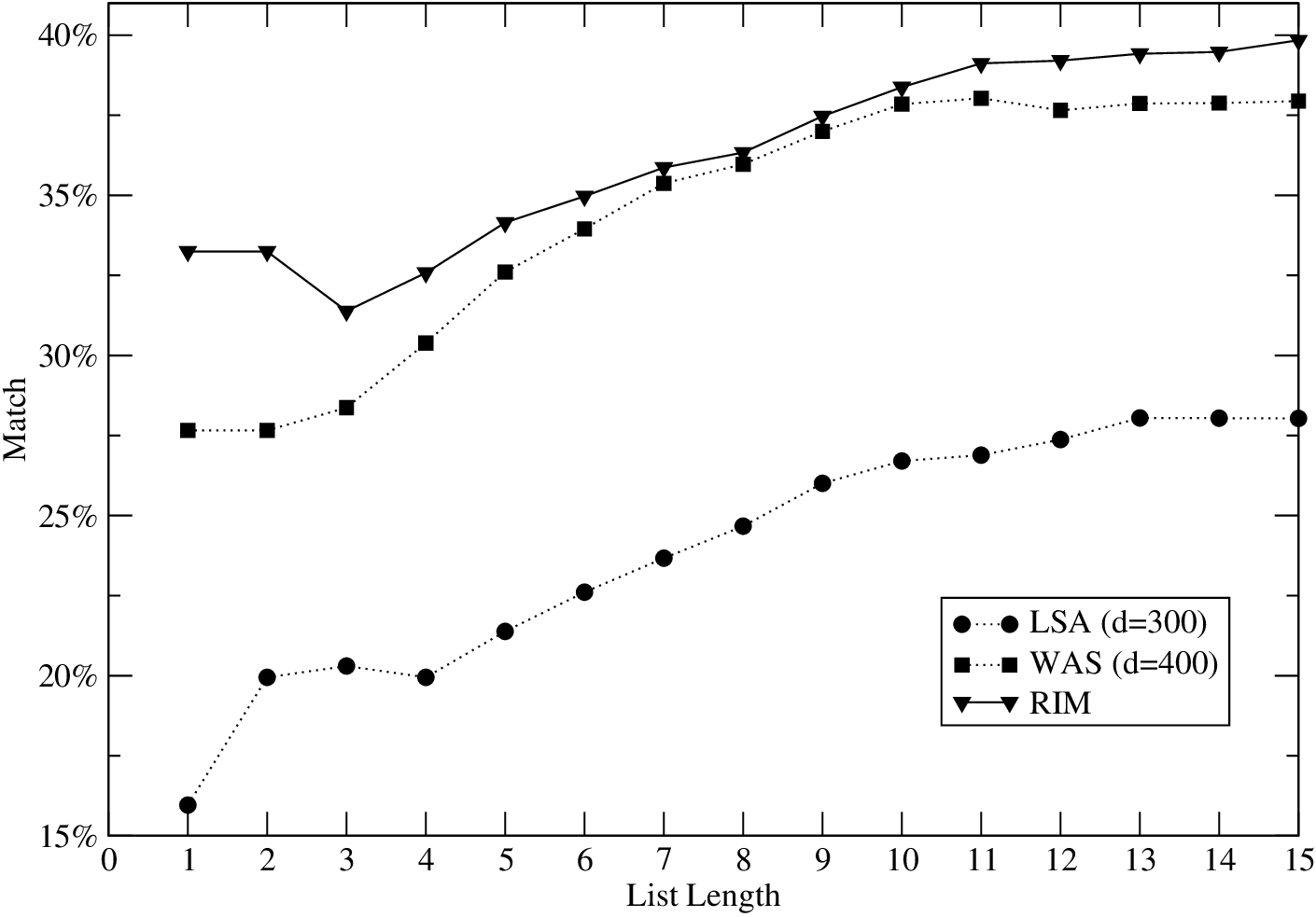}
\caption{Matching average percentage in each synthetic network (LSA, WAS, RIM and RIMCA) against FP. Different vector lengths have been considered ($l=1$ to $l=15$).}
\label{fig:match_data1_10}
\end{figure}

The first problem we face on this microscopic evaluation of the model is that of proposing pertinent measures. We proceed as follows, given a specific word $i$, we start sorting its neighbors according to their linking strength. We apply this for each word in our data sets forming lists. The reference list, is the list of each word in FP, and the lists we want to compare with, are those obtained for each word in the synthetic data sets, RIM, WAS and LSA. We restrict our analysis up to the first 15 ordered neighbors, assuming that these are the most significant ones. We have designed an expression that assigns an error score between a list and its reference, depending on the number of mismatches between both lists, and also on the number of misplacements in them. A mismatch (M) corresponds to a word that exist in the reference list and not in the synthetic list and vice versa, these are considered the main errors in our approach. A misplacement (O) is an error in the order of appearance of both words in each list. The error score E is then defined as:

\begin{equation}
E = E_M + \frac{E_O}{l-E_M}
\label{error_formula}
\end{equation}

\begin{figure}[b]
\centering
\includegraphics[width=0.75\textwidth,clip=] {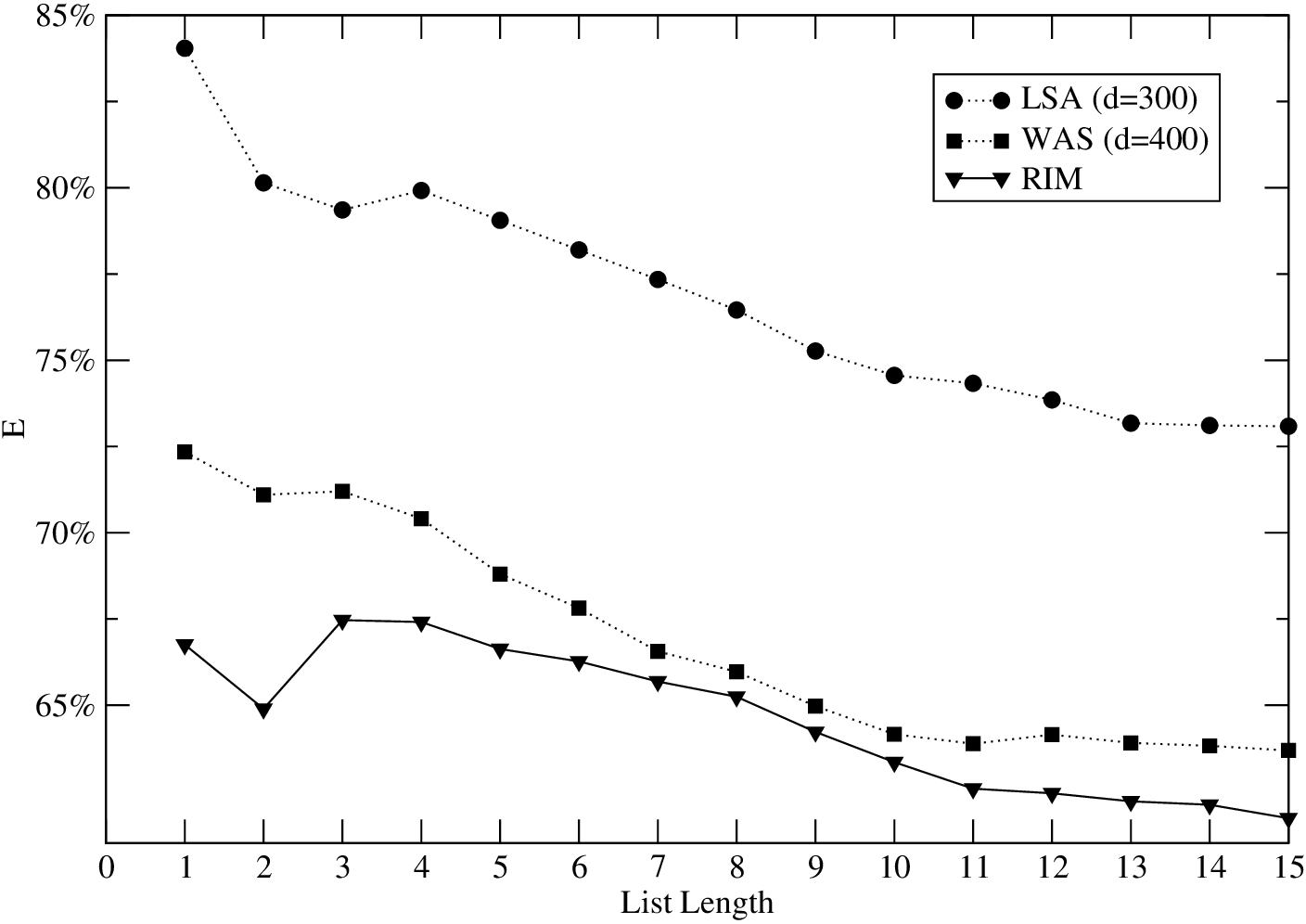}
\caption{For each synthetic network (LSA, WAS and RIM) we have measured the mean error (for $l=1$ to $l=15$) against FP, according to Eq. \ref{error_formula}.}
\label{fig:error_data1_10}
\end{figure}

\noindent where $E_M$ stands for the number of mismatches, $E_O$ the number of displacements and $l$ the length of the list. 
This quantity recalls well-known \emph{edit distances} such as Levenshtein Distance \cite{levenshtein_1} or its generalization, Damerau-Levenshtein Distance \cite{levenshtein_2}, where the similarity between two strings depends on the amount of insertions/deletions (mismatches in our case) and transpositions (movements) that one has to perform on a string in order to completely match the other one. Notice that measure $E$ is strongly increased when a mismatch appears, whereas movements are less punished. In particular, $E = 0$ when the vectors under comparison are identical. In the other extreme, if a vector has only one match with the other one, and the matching element is not placed correctly, then $E = l$, where $l$ is the length of the involved vectors. Beyond this, there exists only a worse situation, i.e. complete mismatch between vectors. In this case $E = \infty$. Since we intend to compute an average error score, we can not allow an $\infty$ value, and therefore we prescribe $E = l+1$, expressing the fact that such score is worse than any case where any match occurs.

In Figure \ref{fig:match_data1_10} we present the frequency of success, proportion of matches in lists of length from 1 to 15, obtained by LSA and RIM. The average matches on these lists is around 24\% in LSA, 34\% in WAS and 36\% in RIM. The error defined in formula \ref{error_formula} is plotted in Figure \ref{fig:error_data1_10}, on average the error of RIM is about 10\% lower than the error of LSA, and 4\% lower than that of WAS. 
An example of the ordered lists of neighbors obtained by models is presented in Table \ref{tab:l10_list} ($l=10$), boldface types are used to highlight the coincidence with the empirical FP.   

\begin{table}[]
\caption{Some illustrative examples of LSA, WAS and RIM's predictive capacity, when compared to our FP ($l=10$ vectors).}
\begin{tabular}{cccc}
TUBA&&&\\
\small{FP}&\small{LSA}&\small{WAS}&\small{RIM}\\\hline
trombone&\bf{clarinet}&bathtub&\bf{trombone}\\
trumpet&violin&faucet&\bf{saxophone}\\
drum&\bf{flute}&sink&\bf{trumpet}\\
cello&guitar&bucket&\bf{flute}\\
clarinet&\bf{trombone}&bridge&\bf{clarinet}\\
saxophone&fork&submarine&\bf{cello}\\
flute&\bf{trumpet}&drain&violin\\
harp&cake&raft&\bf{harp}\\
banjo&\bf{drum}&tap&\bf{banjo}\\
piano&\bf{piano}&dishwasher&stereo\\\hline
\small{ERROR}&4.83&11&2.5\\
&&&\\
ROOSTER&&&\\
chicken&cat&\bf{chicken}&\bf{chicken}\\
goose&gate&crow&\bf{turkey}\\
pigeon&donkey&skillet&crow\\
sparrow&barn&rice&robin\\
penguin&turnip&spinach&\bf{sparrow}\\
pelican&owl&bowl&\bf{bluejay}\\
bluejay&pig&beans&\bf{pigeon}\\
dove&fence&robin&\bf{pelican}\\
hawk&lion&tomato&\bf{goose}\\
turkey&strawberry&\bf{sparrow}&\bf{hawk}\\\hline
\small{ERROR}&11&8.5&2.87\\
\end{tabular}
\label{tab:l10_list}
\end{table}

\section{Conclusions}

In this work, we have proposed an algorithm to extract feature similarity information from an empirical words' Free Association network. Building upon the idea that free association entangles, in particular, semantic traits of association based on similar characteristics between concepts, we have proposed a simple algorithm to disentangle this information. The results reproduce to a large extent the findings in an empirical Feature Production norms network. The simple strategy of a random navigation process of the actual FA topology and a reinforcement inheritance mechanism suffice to produce relationships comparable to those experimentally obtained. 

The comparison with the powerful LSA and WAS models is indicative of the level of macroscopic and microscopic success of our proposal, notwithstanding the fact that both these models provide useful semantic spaces, from a theoretical and an empirical point of view. Furthermore, beyond the level of success of any of these models, we propose that RIM is an approach that enriches other existing models, in the sense that it introduces a dynamical perspective to the formation of semantic spaces. The random navigation mechanism introduced, far from been an optimal strategy in the search space, ought to start exploring dynamic approaches to the problem of semantic cognition.

Assuming that Free Association semantic networks are good exposures of human semantic knowledge, we speculate that some cognitive tasks can rely on a specific navigation of this network, in particular a simple navigation mechanism based on randomness, structure of the network and reinforcement could be enough to reproduce non trivial relationships of feature similarity between concepts represented as words. Moreover, explicit metadata associated to semantic structural patterns seem to play an important role on information recovery, that could be extended to other cognitive tasks. Given the already detected importance of modular structure in the study of semantic representation (see Topic Model \cite{griffiths1}) we think that disambiguation is perhaps the next affordable challenge along this line of research.

\vspace{0.5cm}

\noindent {\bf Acknowledgments} We thank T. L. Griffiths, M. Steyvers, G. Zamora and S. G\'omez, for helpful comments. This work has been supported by the Spanish DGICYT Project FIS2006-13321-C02-02.

\bibliography{disentanglement}

\end{document}